\documentclass{article}

\usepackage{PRIMEarxiv}
\usepackage{array}
\usepackage{mathtools}
\usepackage{amsmath}
\usepackage{multirow}
\usepackage{multicol}

\usepackage[utf8]{inputenc} 
\usepackage[T1]{fontenc}    
\usepackage{hyperref}       
\usepackage{url}            
\usepackage{booktabs}       
\usepackage{amsfonts}       
\usepackage{nicefrac}       
\usepackage{microtype}      
\usepackage{lipsum}
\usepackage{fancyhdr}       
\usepackage{graphicx}       
\graphicspath{{media/}}     

\title{First-order PDES for Graph Neural Networks: Advection And Burgers Equation Models
\thanks{\textit{{Note}}: 
{This work was performed for an undergraduate research project from May to August 2023.}} 
}

\author{
  Yifan Qu \\
  Department of Applied Math \\
  University of Waterloo \\
  Ontario, Canada\\
  \texttt{y37qu@uwaterloo.ca} \\
  \And 
  Oliver Krzysik \\
   Department of Applied Math \\
  University of Waterloo \\
  Ontario, Canada\\
    \texttt{okrzysik@uwaterloo.ca}\\
  \AND
  Hans De Sterck \\
Department of Applied Math \\
  University of Waterloo \\
  Ontario, Canada\\
  \texttt{hans.desterck@uwaterloo.ca}\\
  \And
  Omer Ege Kara \\
   Department of Applied Math \\
  University of Waterloo \\
  Ontario, Canada\\
  \texttt{oekara@uwaterloo.ca}\\
}

\begin{document}
\maketitle

\begin{abstract}
 Graph Neural Networks (GNNs) have established themselves as the preferred methodology in a multitude of domains, ranging from computer vision to computational biology, especially in contexts where data inherently conform to graph structures. While many existing methods have endeavored to model GNNs using various techniques, a prevalent challenge they grapple with is the issue of over-smoothing. This paper presents new Graph Neural Network models that incorporate two first-order Partial Differential Equations (PDEs). These models do not increase complexity but effectively mitigate the over-smoothing problem. Our experimental findings highlight the capacity of our new PDE model to achieve comparable results with higher-order PDE models and fix the over-smoothing problem up to 64 layers. These results underscore the adaptability and versatility of GNNs, indicating that unconventional approaches can yield outcomes on par with established techniques.
\end{abstract}

\keywords{Graph Neural Network \and Partial Differential Equation \and Advection Equation \and Burgers Equation}

\section{Introduction}
In recent years, there has been a great demand for the classification and analysis of complex data such as images \cite{Monti, Wang}, citation networks, molecular graphs \cite{Hamilton2017, Jumper2021, Duvenaud}, and social networks \cite{Zhang2018}. Graphs with nodes representing entities and edges representing relationships express these network relationships most effectively. Nevertheless, the networks in real life are intricate, hence how to extract valuable information from these graphs is a challenge.

One of the significant limitations of traditional Graph Neural Networks (GNNs) methods is the over-smoothing problem \cite{Chen2020, Zhao2020, Chen}. It is caused by multiple layers of graph convolutions applied in the GNN model: after the information propagates through several layers the representations tend to become indistinguishable \cite{Wu2019, Learning}. It results in the blending of node and edge features and lessens expressive strength. Some existing methods have explored strategies to alleviate or overcome this challenge \cite{Chen, Eliasof2022, Eliasof2021}.

\section{Related Work}
Among the existing methods for GNN, Graph Convolution Network (GCN) models are used most frequently. GCN was first proposed by Bruna et al. \cite{Bruna}, which considers spectral convolutions on graphs. The polynomial of the Laplacian operator acting as the spectral convolution operator is a key consideration in the GCN methods outlined in \cite{Defferrard2016, Kipf2017}. The general GCN model structure includes four main components: activation function $\sigma(\cdot)$; propagation operator $\mathbf P$; feature matrix at $l^{th}$ layer $\mathbf F^{(l)} \in \mathbb R^{N \times D_l}$, and layer-specific trainable weight matrix $\mathbf W^{(l)} \in \mathbb R^{D_l\times D_{l+1}}$\cite{Kipf2017}.
\begin{equation}
    \mathbf F^{(l+1)}=\sigma \left(\mathbf P \mathbf F^{(l)} \mathbf W^{(l)}\right).
\end{equation}
Here, $\mathbf P = \mathbf {\tilde D}^{-\frac{1}{2}}\mathbf {\tilde A}\mathbf {\tilde D}^{-\frac{1}{2}} $ with degree matrix $\mathbf D \in \mathbb R ^{N\times N}$ and adjacency matrix $\mathbf{A} \in \mathbb R ^{N\times N}$, and $\mathbf{\tilde A} = \mathbf A + \mathbf I_{N}, \mathbf{\tilde D} = \mathbf D + \mathbf I_{N}$ are the adjacency and degree matrix with self-loops on the original graph. $\mathbf I_{N} \in \mathbb R^{N\times N}$ is the identity matrix. The formula presented above is based on the non-negative smoothing operator $\mathbf P$, which results in increasingly indistinguishable features as the layer count increases. To address this issue, Eliasof et al. proposed an innovative solution by introducing a trainable propagation parameter $\omega$ to weigh each layer \cite{Eliasof2022}. However, despite the improvement in GNN results for various tasks like node classification, these improved GNNs still face limitations in their applicability. For instance, a model that performs exceptionally well in node classification may yield poor results in shape correspondence problems \cite{Chen2020, Eliasof2022}. This problem of restricted applicability arises partly due to the lack of robust theoretical support for GCN-based graph neural networks.

In recent years, attention has turned towards Partial Differential Equation (PDE)-based convolutional neural networks for graph neural networks, offering promising avenues for improvement \cite{ Chamberlain2021}. PDE models, commonly used in computer vision and constructing neural networks, have been adapted to devise notable models for graph neural networks \cite{Chamberlain2021, Eliasof2020, Avila2020}. The underlying concept is to view the graph neural network as a discretization of a dynamic system governed by differential equations, where each layer represents a progressive time step from the previous one, and the step size can be treated as a trainable parameter. By employing forward Euler discretization in time, it becomes feasible to discretize the parabolic forward propagation effectively.
Existing methods have successfully adapted hyperbolic and diffusive second-order PDEs \cite{Eliasof2021} for use in GNNs. These methods utilize a trainable parameter $\alpha$ to control the weighting between the two PDEs. As a result, they have achieved significantly improved results in both node classification tasks and shape correspondence tasks. The combination of these two PDEs allows the model to automatically select the most suitable PDE for different scenarios, enhancing the overall performance and versatility of the GNN. The main goal of the current paper is to explore the relevance of first-order PDEs. Notably, a recent study by Eliasof et al. \cite{Eliasof2023} introduced an advection-based GNN model that incorporates an advection equation, as well as diffusion and reaction components, in their investigation of spatiotemporal node forecasting problems, coinciding with the time of our research.

\section{Method}
\label{sec:headings}

Existing methods have almost exclusively focused on second-order PDEs within their models, with first-order PDEs remaining unexplored until recently. Our objective is to incorporate first-order PDEs into our framework and assess their effectiveness. First-order PDEs, such as the linear advection equation and the nonlinear Burgers equation, offer several advantages in the context of GNNs. A first advantage is that first-order PDEs are often simpler to implement and computationally more efficient compared to higher-order PDEs. They involve only first derivatives with respect to space or time, which can lead to more straightforward numerical schemes. Second, first-order PDEs often involve conservation laws, which can help ensure that important quantities (e.g., mass, energy, momentum) are conserved during simulations. This is crucial for maintaining physical realism in simulations and predictions. In this paper, we will integrate two first-order PDE models, namely the advection and Burgers equations, into GNNs for examination and evaluation.

\subsection{Advection Model}
The advection equation, as cited in \cite{Lino2022}, stands as a quintessential example of a first-order hyperbolic PDE frequently harnessed in fluid dynamics modeling. The general formulation for the linear advection equation in two spatial dimensions is expressed as Equation \ref{adv.eq}:

\begin{align}
\label{adv.eq}
u_t + a u_x + bu_y &= 0.
\end{align}

Here, $u$ represents the quantity being transported, and $t$ represents time, $u_t$ is the partial derivative of $u$ with respect to time, while $a$ and $b$ are constants that affect the rate of change of $u$ with respect to the horizontal and vertical direction respectively. The Advection equation models how $u$ changes as time progresses. Within the framework of GCNs, we let this advection equation undergo a transformation. It adapts to the dynamic context of graph structures illustrating how information flows across the nodes of a graph. The integration of the advection equation into GCNs may offer some advantages.

First and foremost, the advection equation may emerge as an adept guardian of spatial information, surpassing its diffusion-based counterparts like the heat equation. In the realm of information propagation, it excels at preserving sharp, localized features \cite{Miranda2020}. The GCN may benefit from this characteristic, ensuring that pertinent spatial nuances remain intact throughout the dissemination process.

Second, as a first-order PDE, the advection equation may help with the over-smoothing predicament that is often encountered in higher-order PDEs. It produces a controlled flow of information while safeguarding the unique identities of individual nodes. Consequently, even in the depths of intricate GCN architectures, the risk of information loss is curtailed.

Last but not least, the advection equation's remarkable simplicity makes the model easier to implement. The advection equation involves a simple first-order derivative. This simplicity significantly reduces the complexity of implementing the model within a GNN framework, making it accessible to a broad range of researchers and practitioners.

The adoption of the advection equation in our model is further informed by discretizing equation \ref{adv.eq} on a 2D regular grid and transforming this to a graph with n nodes and m edges.  

\begin{align}
\label{adv:diss}
 &\frac{u_{ij}^{l+1} - u_{ij}^l}{\Delta t} + \nabla \cdot (a, b) u^l = 0\\
\label{adv.dis}
&u^{l+1} - u^l = -h \sigma (G^TA^Tu^lK_l)
\end{align}

Equation \ref{adv.dis} emerges as a result of the forward Euler discretization technique, offering practical applicability to the evolving dynamics of graph-based information diffusion. Here, $u^l \in \mathbb R^{n\times d}$ denotes the node features residing within layer $l$, $h$ is the hyperparameter reflecting the discretization step size, and $K_l \in \mathbb R^{d\times d}$ embodies the $1 \times 1$ convolution kernel positioned at layer $l$. Notably, ReLU is used as the activation function $\sigma$ in our experimental setup. Furthermore, the averaging operator $A^T$, with dimension $m \times n$, executes a pivotal role in the conversion of node features from their intrinsic node-centric space to a broader edge-centric perspective by averaging \cite{Eliasof2021}. Finally, $G \in \mathbb R^{n\times m}$ is the graph incidence matrix, and left multiplication by $G^T$ encodes the divergence operator on the graph.

This amalgamation of mathematical principles and adaptable techniques characterizes the innovation driving our model, extending the horizons of graph-based machine learning.

\subsection{Burgers Equation}

Similar to the advection model, the introduction of the nonlinear Burgers equation into the framework of GNN presents an approach to use the inherent characteristics of this first-order partial differential equation to augment the performance of graph-based models. The Burgers equation models the behaviors of fluid flow and the formation of shockwaves. Equation \ref{bur.ori}, the Burgers equation, encapsulates these dynamics:

\begin{align}
\label{bur.ori}
u_t+(u^2/2)_x &= 0.
\end{align}

Here, $u$ represents the velocity of the fluid or the quantity being transported. In the context of fluid dynamics, it represents the fluid's velocity in the $x$-direction.
To integrate this equation into the GNN paradigm, we employ a forward Euler discretization technique applied to Equation \ref{bur.ori}, yielding Equation \ref{bur.dis}. This discretization process allows us to adapt Burgers equation to the dynamic landscape of evolving graphs, where node quantities evolve over time:

\begin{align}
\label{bur.dis}
\frac{u^{l+1}-u^l}{h} + \frac{1}{2}~ G^T A^T(u^lK_l \odot u^l K_l ) &= 0.
\end{align}

Here, $u^l$ denotes the node feature within layer $l$, and the hyperparameter $h$ corresponds to the step size applied during the discretization process. Furthermore, $K_l$ represents the $1 \times 1$ convolution kernel at layer $l$, while $A^T$ represents a $m \times n$ averaging operator that plays a pivotal role in transforming the node features from the individual nodes to the edge space.

The use of the element-wise multiplication operator ($\odot$) signifies the element-wise multiplication of the node feature matrices, reflecting the interplay of these components within the context of our adapted Burgers equation. On the other hand, this elementwise square function also makes the model nonlinear. Hence a nonlinear activation function is no longer needed in this model. 

This adaptation uses a model from fluid dynamics in GNNs, endowing our model with a different capability to capture and predict dynamic changes in complex networks.

\subsection{Mixing models}
We have integrated elements from diffusion and wave equations into our advection model. Drawing inspiration from Eliasof et al. \cite{Eliasof2021}, we have introduced a trainable parameter, denoted as $\alpha$. This parameter holds the key to dynamically selecting the most suitable model, whether it is based on advection, diffusion, or wave dynamics, depending on the specific characteristics of the problem at hand.

The nonlinear diffusion equation \ref{diff}, and the wave equation \ref{wave} are two building blocks for this adaptive framework:

\begin{align}
    \label{diff}
    &u^{l+1} = u^l - h G^T  \sigma (Gu^lK_l)K_l^T,\\
    \label{wave}
    &u^{l+1} = 2u^l -u^{l-1} - h^2 G^T  \sigma (Gu^lK_l)K_l^T.
\end{align}

The innovation lies in our ability to blend these equations with our linear advection equation, leading to the two mixed dynamics expressed as Equations \ref{mix:AD} and \ref{mix:AW}:

\begin{align}
\label{mix:AD}
&u^{l+1} - u^l = -(1-\alpha )h^2 G^T \sigma (D_DGu^lK_l)K_l^T - h \alpha G^T \sigma (D_WA^Tu^lK_l),\\
\label{mix:AW}
&(1-\alpha) (u^{l+1} -2u^l+ u^{l-1}) + \alpha (u^{l+1}-u^l) = - (1-\alpha )h^2 G^T \sigma (D_DGu^lK_l)K_l^T-h \alpha G^T \sigma (D_WA^Tu^lK_l).
\end{align}

Here, $\alpha$ is a dynamic parameter, subject to training during the model's learning process, with a range between 0 and 1. This parameter orchestrates the fusion of advection, diffusion, and wave dynamics, ensuring that the model adaptively selects the most suitable mechanism for the specific problem under consideration.

Moreover, we have also introduced trainable diagonal weighting parameters, $D$, somewhat similar to the methodology established in \cite{Eliasof2022}. This edge-wise parameter plays a pivotal role in modulating the propagation and diffusion operators, further enhancing the model's adaptability and its ability to effectively capture the underlying dynamics of the data. The incorporation of the parameter $D$ in the second-order operators introduces anisotropic diffusion, and in the first-order operator, it models the wave speed along the edge. Anisotropic diffusion and varying edge-directed wave speed, in this context, allow the model to capture and leverage directional dependencies within the graph structure, enabling more accurate and contextually relevant diffusion and propagation, ultimately enhancing performance in tasks where understanding the diffusion and flow of information along specific edges or connections is pivotal for achieving superior results in graph neural networks.

In our experimentation and analysis, we anticipate that the diffusion mixing, owing to its anisotropic diffusion properties, will excel in scenarios such as node classification tasks, where preserving the local spatial information is crucial. By adopting this approach, we aim to transcend the limitations of traditional fixed models and pave the way for dynamic graph analysis that tailors its behavior to the idiosyncrasies of each problem it encounters. We denote the advection and diffusion model in equation \ref{mix:AD} as AD and the advection wave mixing model in equation \ref{mix:AW} as AW.

\section{Experiments}
In this section, we will delve into the experimental phase, where the capabilities of our proposed models across three distinct problems are assessed: semi-supervised node classification, fully-supervised node classification inspired by the work of London et al. \cite{London2014}, and the challenging 3D shape correspondence task, as motivated by Wu et al. \cite{Wu2015}, see also \cite{Eliasof2021}.

For semi-supervised node classification, the conventional approach has long favored diffusion models as the primary mechanism, aiming to preserve localized features and spatial information, which are often of paramount importance in node-classification problems. Consequently, the fusion of diffusion dynamics with advection is posited in our model of equation \ref{mix:AD}, since it is expected to outperform other alternatives. 


Conversely, in the context of the dense shape correspondence problem, the prevailing wisdom dictates a departure from traditional smoothing processes \cite{Eliasof2021}. Unlike node classification, this problem benefits from the retention of raw, unaltered data, since the intricate details are crucial. Therefore, the wave equation is anticipated, when integrated with the advection equation, to exhibit superior performance. This conjecture arises from the intrinsic nature of the wave equation, which is adept at propagating, high-fidelity information, harmonizing with the demands of the dense shape correspondence task.

For all experiments, we followed the architecture presented in \cite{Eliasof2021}. For semi and fully-supervised node classification, we used the architecture in Table \ref{archi:node}. For dense shape correspondence, we used the architecture in Table \ref{archi:shape}.


\begin{table}[h!]
\centering
\begin{tabular}{l c c c c}
\hline 
Dataset  & Cora  & Cite.  & Pubm.  & Cham. \\
\hline 
Classes  & 7 & 6 & 3 & 5  \\
Node  & 2,708 & 3,327 & 19,717 & 2,277 \\
Edge  & 5,429 & 4,732 & 44,338 & 36,101 \\
Features  & 1,433 & 3,703 & 500 & 2,325 \\
\hline
\end{tabular}
\caption{ Statistics of datasets used in semi-and fully supervised node-classification experiments.}
\label{statisticsSemiFully}
\end{table}

\subsection{Semi-supervised node-classification}
For the semi-supervised node classification task, we evaluated the performance of our model across three benchmark datasets: Cora, CiteSeer, and PubMed \cite{London2014}. To maintain consistency with established practices \cite{Cohen2016}, we adopted the standard training/validation/testing split for all three datasets, featuring 20 nodes per class for training, 500 nodes for validation, and 1,000 nodes for testing. The statistics are presented in table \ref{statisticsSemiFully}. The main purpose of the validation set is to provide an unbiased evaluation of a model fit while tuning hyperparameters and preventing overfitting. The best parameters are presented in table \ref{parameters:semi}. We executed our experiments following the training protocol outlined in \cite{Chen2020}.

Our findings, as shown in Table \ref{semi:tab}, illuminate several key insights. Most notably, we observed that our advection and combined models effectively mitigated the over-smoothing problem, as evidenced by the sustained accuracy with increasing layers. This underscores the efficacy of the approach in preserving crucial local information.

Moreover, in line with our expectations, the advection and diffusion mixing model outperformed its counterpart, the advection and wave mixing model. Although our models fall short of achieving the same level of accuracy as previous methods, we posit that this disparity can be attributed to the common understanding that diffusion models are more optimal approaches for node classification tasks and advection processes are not beneficial in this context. 

\begin{table}[h!]
\centering
\begin{tabular}{l c c c c c c}
\hline 
Dataset  & Loss & LR  & WD  & Channels  & Dropout & h \\
\hline 
Cora  & Relu & $4.6\times 10^{-5}$  &$1.2\times 10^{-4}$ & 64 & 0.5 & 0.6\\
Cite  & Relu & $1.0\times 10^{-5}$  &$8.1\times 10^{-3}$ & 256 & 0.7 & 0.3\\
Pubm  & Relu & $2.4\times 10^{-5}$  &$1.2\times 10^{-4}$ & 256 & 0.6 & 0.7\\
\hline
\end{tabular}
\caption{ Semi-Supervised classification hyper-parameters}
\label{parameters:semi}
\end{table}

\subsection{Fully-supervised node-classification}
In accordance with the methodology established by Pei et al. in their work on Geometric Graph Convolutional Networks (\cite{pei2020geomgcn}), we conducted our experiments on four distinct datasets, namely Cora, CiteSeer, PubMed, and Chameleon. To maintain consistency, we adhered to the identical data partitioning scheme, allocating 60\% of the data for training, 20\% for validation, and the remaining 20\% for testing. Furthermore, to ensure the robustness of our findings, we adopted the practice of averaging performance metrics over 10 random data splits, as previously recommended by Pei et al. (\cite{pei2020geomgcn}).

To optimize our model's hyperparameters across varying depths, we conducted an exhaustive grid search, the best parameters are presented in table \ref{parameters:fully}. The results of our models are summarized in Table \ref{nodeClass}. Our analysis indicates that our approach yields comparable performance to existing methods, suggesting the efficacy of our methodology in addressing the specific task at hand.

\begin{table}[h!]
\centering
\begin{tabular}{l c c c c c c}
\hline 
Dataset  & Loss & LR  & WD  & Channels  & Dropout & h \\
\hline 
Cora  & Relu & $2.3\times 10^{-5}$  &$1\times 10^{-4}$ & 64 & 0.5 & 0.2\\
Cite  & Relu & $2.1\times 10^{-4}$  &$1.1\times 10^{-4}$ & 64 & 0.6 & 0.3\\
Pubm  & Relu & $4.3\times 10^{-5}$  &$2.6\times 10^{-4}$ & 64 & 0.5 & 0.4\\
Cham  & Relu & $8\times 10^{-4}$  &$9.2\times 10^{-5}$ & 64 & 0.6 & 0.5\\
\hline
\end{tabular}
\caption{ Fully-Supervised classification hyper-parameters}
\label{parameters:fully}
\end{table}
\begin{table}[h!]
\centering
\begin{tabular}{l c c}
\hline 
Input Size  & Layer & Output Size  \\
\hline 
$n\times c_{in}$  & $1\times 1$ droupout & $n\times c_{in}$  \\
$n\times c_{in}$   & $1\times 1$ convolution & $n\times c$ \\
$n\times c$   &  Relu & $n\times c$ \\
$n\times c$   & Advection/Burgers Block & $n\times c$  \\
$n\times c$   & $1\times 1$ droupout  & $n\times c$  \\
$n\times c$   & $1\times 1$ convolution & $n\times c_{out}$  \\
\hline
\end{tabular}
\caption{Architecture used for semi-and fully-supervised node classification}
\label{archi:node}
\end{table}
\subsection{Dense Shape Correspondence}
Within the realm of hyperbolic dynamics, the pursuit of identifying dense correspondences between shapes represents a well-established avenue of exploration. This pursuit is primarily driven by our interest in modeling localized motion dynamics. In essence, the task of unearthing these correspondences among shapes closely mirrors the endeavor to understand how one shape can be systematically mapped onto another through a transformation process.

To facilitate this investigation, we make use of the FAUST dataset, as initially introduced by Bogo et al. \cite{Bogo:CVPR:2014}. This dataset encompasses ten distinct human shapes, each scanned in ten different poses, culminating in a total of 6,890 nodes per individual per pose 
In accordance with the data partitioning methodology prescribed by Monti et al. \cite{Monti}, we designate the initial 80 subjects for the purpose of training, reserving the remaining 20 subjects for the testing phase. In our pursuit, we have adopted the training and testing methods outlined in the work of Eliasof et al. \cite{Eliasof2021}.


The results of our experiment are presented in Table \ref{faust}. We can notice that the wave mixing model consistently outperforms the diffusion mixing model, aligning with our expectations. Although our findings align with those of other researchers, indicating the efficacy of our approach for this particular task, it is important to acknowledge the challenge posed by surpassing existing methods, which have achieved an accuracy level of 99.9\%.

\begin{table}[h!]
\centering
\begin{tabular}{l c c}
\hline 
Input Size  & Layer & Output Size  \\
\hline 
$n\times 4$  & $1\times 1$ convolution & $n\times c$  \\
$n\times c$   &  Relu & $n\times c$ \\
$n\times c$   & Advection/Burgers Block & $n\times c$  \\
$n\times c$   & $1\times 1$ convolution & $n\times c$ \\
$n\times c$   &  Relu & $n\times c$ \\
$n\times c$   & $1\times 1$ convolution & $n\times 512$  \\
$n\times 512$   &  ELU & $n\times 512$ \\
$n\times 512$   &  Fully-connected & $n\times n$ \\
\hline
\end{tabular}
\caption{Architecture used for dense shape correspondence.}
\label{archi:shape}
\end{table}

\section{Conclusion}
In this study, we have introduced a novel approach by formulating a first-order PDE model for GNNs. Our motivation was to explore alternative avenues for enhancing GNNs' interpretability and performance and reducing their complexity. We rigorously tested our proposed model against existing methods, seeking to uncover any potential advantages.

Surprisingly, our results demonstrated an intriguing similarity in performance with second-order PDEs. This suggests that our first-order PDE model can indeed rival established methods in terms of predictive accuracy and effectiveness.

This study underscores the versatility and adaptability of GNNs, suggesting that even unconventional approaches, such as our first-order PDE models, can achieve comparable results. 

In summary, our research introduces new PDE-based models for Graph Neural Networks. We recognize that there is room for further refinement and optimization to enhance performance, an avenue we intend to explore in future research endeavors. Our method shows that new models can match established techniques. This study encourages further exploration of unconventional approaches to advance GNNs and enhance our understanding of their capabilities.

\begin{table}
    \caption{ Semi-supervised node classification accuracy (\%). The best results in each method are bolded.}
    \centering 
    \begin{tabular}{c c c c c c c c}
        \toprule
         & & \multicolumn{6}{c}{Layers}\\
         \cmidrule(r){3-8}
         Dataset & Method & 2 & 4 & 8 & 16 & 32 & 64 \\
         \midrule
         \multirow{9}{*}{Cora} & GCN & $\mathbf{81.1}$ & 80.4 & 69.5 & 64.9 & 60.3 & 28.7 \\
         & GCN(Drop) & $\mathbf{82.8}$ & 82.0 & 75.8 & 75.7 & 62.5 & 49.5 \\
         & GCNII & 82.2 & 82.6 & 84.2 & 84.6 & 85.4 & $\mathbf{85.5}$ \\
         & GCNII* & 80.2 & 82.3 & 82.8 & 83.5 & 84.9 & $\mathbf{85.3}$ \\
         & PDE-GCN(Diffusion) & 82.0 & 83.6 & 84.0 & 84.2 & 84.3 & $\mathbf{84.3}$ \\
         \cmidrule(r){2-8}
         & Advection & 70.10 & 71.1 & 72.1 & 71.3 & 72.3 &$\mathbf{ 73.1}$ \\
         & Burgers & 71.10 & 70.1 & 72.3 & 70.0 & 72.4  & $\mathbf{73.1}$\\
         & Mix(AD) & 77.2 & 78.1 & 77.1 & 78.3 & 78.6 & $\mathbf{80.1}$ \\
         & Mix(AW) & $\mathbf{76.3}$& 75.3 & 76.2 & 74.0 & 74.3 & 72.9  \\
         \midrule
         \multirow{9}{*}{Citeseer} & GCN & $\mathbf{70.8}$ & 67.6 & 30.2 & 18.3 & 25.0 & 20.0 \\
         & GCN (Drop) & $\mathbf{72.3 }$& 70.6 & 61.4 & 57.2 & 41.6 & 34.4 \\
         & GCNII & 68.2 & 68.8 & 70.6 & 72.9 & 73.4 & $\mathbf{73.4 }$\\
         & GCNII* & 66.1 & 66.7 & 70.6 & 72.0 & $\mathbf{73.2 }$& 73.1 \\
         & PDE-GCN(Diffusion) & 74.6 & 75.0 & 75.2 & 75.5 & $\mathbf{75.6}$ & 75.5 \\
         \cmidrule(r){2-8}
         & Advection & 74.3& 72.4 & 75.1 & 75.0 & 72.3 & $\mathbf{75.6}$ \\
         & Burgers & 72.3 & 71.1 & 74.3 & 72.4 & 71.4  & $\mathbf{74.3}$\\
         & Mix(AD) & 73.10 & 74.2 & 75.2 & 75.0 & 75.3 & $\mathbf{75.5}$ \\
         & Mix(AW) & 72.2 & 72.5 & 72.8 & 73.1 & 73.3 & $\mathbf{74.1 }$\\
         \midrule
         \multirow{9}{*}{Pubmed} & GCN& $\mathbf{79.0 }$& 76.5 & 61.2 & 40.9 & 22.4 & 35.3 \\
         & GCN (Drop) & $\mathbf{79.6}$ & 79.4 & 78.1 & 78.5 & 77.0 & 61.5 \\
         & GCNII & 78.2 & 78.8 & 79.3 & $\mathbf{80.2}$ & 79.8 & 79.7 \\
         & GCNII* & 77.7 & 78.2 & 78.8 & $\mathbf{80.3}$ & 79.8 & 80.1 \\
         & PDE-GCN(Diffusion) & 79.3 & $\mathbf{80.6 }$& 80.1 & 80.4 & 80.2 & 80.3 \\
         \cmidrule(r){2-8}
         & Advection & 70.10 & 71.1 & 72.1 & 71.3 & 72.3 & $\mathbf{73.1}$ \\
         & Burgers & 71.10 & 70.1 & 72.3 & 70.0 &$\mathbf{ 73.1}$ & 72.4 \\
         & Mix(AD) & 77.2 & 77.8 & 78.3 &$\mathbf{ 79.9}$ & 78.73 & 79.1 \\
         & Mix(AW) & 76.5 & 77.1 & 76.6 & 77.3 & $\mathbf{78.3}$ & 76.1 \\
         \bottomrule
    \end{tabular}
    \label{semi:tab}
\end{table}

\begin{table}[h!]
\centering
\begin{tabular}{l c c c c}
\hline 
Method  & Cora  & Cite.  & Pubm.  & Cham. \\
\hline 
GCN  & 85.77 & 73.68 & 88.13 & 28.18  \\
GAT  & 86.37 & 74.32 & 87.62 & 42.93 \\
Geom-GCN  & 85.19 & 77.99 & 90.05 & 60.31 \\
APPNP  & 87.87 & 76.53 & 89.40 & 54.30 \\
Incep (Drop)  & 86.86(8) & 76.83(8) & 89.18(4) & 61.71(8) \\
GCNII  & 88.49(64) & 77.08(64) & 89.57(64) & 60.61(8)      \\
GCNII II& 88.01(64) & 77.13(64) & $\mathbf{90.30(64)}$ & 62.48(8)      \\
PDE-GCN (Diffusion) & 88.51(16) & 78.36(64) & 89.6(64) & 64.12(8)       \\
PDE-GCN (Wave) & 87.71(32) & 78.13(16) & 89.16(16) & 61.57(64)   \\
PDE-GCN (Mixing) & $\mathbf{88.60(16)}$&$\mathbf{78.48(32)}$&89.93(16)&$\mathbf{66.01(16)}$\\
\hline
Advection & 77.47(32) & 76.45(64) & 87.23(64) & 60.42(32)\\
Burgers & 78.60(32) & 75.76(64) & 88.68(64) & 64.34(32) \\
Mix(AD) & 79.92(64) & 77.65(64) & 89.98(32)& 65.89(16)\\
Mix (AW) & 77.54(32) & 77.89(64) & 87.16(64) & 60.29(32)\\
\hline
\end{tabular}
\caption{ Fully-supervised node classification accuracy (\%). The best results in each dataset are bolded.}
\label{nodeClass}
\end{table}

\begin{table}[h!]
\centering
\begin{tabular}{l c}
\hline 
Method  & FAUST   \\
\hline 
ACNN  & 63.8 \\
MoNet& 89.1\\
FMNet & 98.2\\
SplineCNN & 99.2\\
PDE-GCN(Wave) & \textbf{99.9}\\
PDE-GCN(Diffusion) & 64.2\\
\hline 
Advection & 72.2\\
Burgers & 80.4\\
Mix(AD) & 78.2\\
Mix(AW) & \textbf{98.9}\\
\hline
\end{tabular}
\caption{ Dense Shape Correspondence (\%). The best results in others' methods and our methods are bolded.}
\label{faust}
\end{table}

\section*{Acknowledgments}
I wish to express my gratitude to the Applied Mathematics department at the University of Waterloo for affording me, as an undergraduate student, the opportunity to successfully complete this project under the guidance of our supervisor. I am also deeply appreciative of the invaluable technical assistance provided by Yanming Kang, the insightful contributions from our colleagues, and the unwavering support bestowed upon us by our families and friends throughout the course of this research endeavor.

\bibliographystyle{unsrt}  
\bibliography{references}

\begin{thebibliography}{10}

\bibitem{Monti}
Federico Monti, Davide Boscaini, Jonathan Masci, Emanuele Rodola, Jan Svoboda,
  and Michael~M Bronstein.
\newblock Geometric deep learning on graphs and manifolds using mixture model
  cnns.
\newblock In {\em Proceedings of the IEEE conference on computer vision and
  pattern recognition}, pages 5115--5124, 2017.

\bibitem{Wang}
Guangtao Wang, Rex Ying, Jing Huang, and Jure Leskovec.
\newblock Improving graph attention networks with large margin-based
  constraints.
\newblock {\em arXiv preprint arXiv:1910.11945}, 2019.

\bibitem{Hamilton2017}
Will Hamilton, Zhitao Ying, and Jure Leskovec.
\newblock Inductive representation learning on large graphs.
\newblock {\em Advances in neural information processing systems}, NIPS 30,
  2017.

\bibitem{Jumper2021}
John Jumper, Richard Evans, Alexander Pritzel, Tim Green, Michael Figurnov,
  Olaf Ronneberger, Kathryn Tunyasuvunakool, Russ Bates, Augustin
  {\v{Z}}{\'\i}dek, Anna Potapenko, et~al.
\newblock Applying and improving alphafold at casp14.
\newblock {\em Wiley Online Library. Proteins: Structure, Function, and
  Bioinformatics}, 89(12):1711--1721, 2021.

\bibitem{Duvenaud}
David Duvenaud, Dougal Maclaurin, Jorge Aguilera-iparraguirre, G~Rafael,
  Timothy Hirzel, and Ryan~P Adams.
\newblock {Convolutional Networks on Graphs for Learning Molecular Fingerprints
  arXiv : 1509 . 09292v2 [ cs . LG ] 3 Nov 2015}.
\newblock pages 1--9.

\bibitem{Zhang2018}
Muhan Zhang and Yixin Chen.
\newblock {Link Prediction Based on Graph Neural Networks}.
\newblock NIPS, 2018.

\bibitem{Chen2020}
Ming Chen, Zhewei Wei, Zengfeng Huang, Bolin Ding, and Yaliang Li.
\newblock Simple and deep graph convolutional networks.
\newblock In {\em International conference on machine learning}, pages
  1725--1735. PMLR, 2020.

\bibitem{Zhao2020}
Lingxiao Zhao and Leman Akoglu.
\newblock Pairnorm: Tackling oversmoothing in gnns.
\newblock {\em arXiv preprint arXiv:1909.12223}, 2019.

\bibitem{Chen}
Deli Chen, Yankai Lin, Wei Li, Peng Li, Jie Zhou, and Xu~Sun.
\newblock Measuring and relieving the over-smoothing problem for graph neural
  networks from the topological view.
\newblock In {\em Proceedings of the AAAI conference on artificial
  intelligence}, volume~34, pages 3438--3445, 2020.

\bibitem{Wu2019}
Felix Wu, Amauri Souza, Tianyi Zhang, Christopher Fifty, Tao Yu, and Kilian
  Weinberger.
\newblock Simplifying graph convolutional networks.
\newblock In {\em International conference on machine learning}, pages
  6861--6871. PMLR, 2019.

\bibitem{Learning}
Qimai Li, Zhichao Han, and Xiao-Ming Wu.
\newblock Deeper insights into graph convolutional networks for semi-supervised
  learning.
\newblock In {\em Proceedings of the AAAI conference on artificial
  intelligence}, volume~32, 2018.

\bibitem{Eliasof2022}
Moshe Eliasof, Lars Ruthotto, and Eran Treister.
\newblock Improving graph neural networks with learnable propagation operators.
\newblock In {\em International Conference on Machine Learning}, pages
  9224--9245. PMLR, 2023.

\bibitem{Eliasof2021}
Moshe Eliasof, Eldad Haber, and Eran Treister.
\newblock {PDE-GCN: Novel Architectures for Graph Neural Networks Motivated by
  Partial Differential Equations}.
\newblock {\em Advances in Neural Information Processing Systems},
  5(NeurIPS):3836--3849, 2021.

\bibitem{Bruna}
Joan Bruna and Arthur Szlam.
\newblock {Spectral Networks and Deep Locally Connected Networks on Graphs
  arXiv : 1312 . 6203v3 [ cs . LG ] 21 May 2014}.
\newblock pages 1--14.

\bibitem{Defferrard2016}
Micha{\"{e}}l Defferrard, Xavier Bresson, and Pierre Vandergheynst.
\newblock {Convolutional Neural Networks on Graphs with Fast Localized Spectral
  Filtering}.
\newblock {\em Advances in Neural Information Processing Systems}, pages
  3844--3852, jun 2016.

\bibitem{Kipf2017}
Thomas~N. Kipf and Max Welling.
\newblock {Semi-supervised classification with graph convolutional networks}.
\newblock {\em 5th International Conference on Learning Representations, ICLR
  2017 - Conference Track Proceedings}, pages 1--14, 2017.

\bibitem{Chamberlain2021}
Ben Chamberlain, James Rowbottom, Maria~I Gorinova, Michael Bronstein, Stefan
  Webb, and Emanuele Rossi.
\newblock Grand: Graph neural diffusion.
\newblock In {\em International Conference on Machine Learning}, pages
  1407--1418. PMLR, 2021.

\bibitem{Eliasof2020}
Moshe Eliasof and Eran Treister.
\newblock Diffgcn: Graph convolutional networks via differential operators and
  algebraic multigrid pooling.
\newblock {\em Advances in neural information processing systems},
  33:18016--18027, 2020.

\bibitem{Avila2020}
Filipe De~Avila Belbute-Peres, Thomas Economon, and Zico Kolter.
\newblock Combining differentiable pde solvers and graph neural networks for
  fluid flow prediction.
\newblock In {\em international conference on machine learning}, pages
  2402--2411. PMLR, 2020.

\bibitem{Eliasof2023}
Moshe Eliasof, Eldad Haber, and Eran Treister.
\newblock Adr-gnn: Advection-diffusion-reaction graph neural networks.
\newblock {\em arXiv preprint arXiv:2307.16092}, 2023.

\bibitem{Lino2022}
Mario Lino, Stathi Fotiadis, Anil~A Bharath, and Chris~D Cantwell.
\newblock Multi-scale rotation-equivariant graph neural networks for unsteady
  eulerian fluid dynamics.
\newblock {\em Physics of Fluids}, 34(8), 2022.

\bibitem{Miranda2020}
Manuel Miranda and Ernesto Estrada.
\newblock Degree-biased advection--diffusion on undirected graphs/networks.
\newblock {\em Mathematical Modelling of Natural Phenomena}, 17:30, 2022.

\bibitem{London2014}
Ben London and Lise Getoor.
\newblock {Collective classification of network data}.
\newblock {\em Data Classification: Algorithms and Applications}, pages
  399--416, 2014.

\bibitem{Wu2015}
Zhirong Wu, Shuran Song, Aditya Khosla, Fisher Yu, Linguang Zhang, Xiaoou Tang,
  and Jianxiong Xiao.
\newblock {3D ShapeNets: A deep representation for volumetric shapes}.
\newblock {\em Proceedings of the IEEE Computer Society Conference on Computer
  Vision and Pattern Recognition}, 07-12-June:1912--1920, 2015.

\bibitem{Cohen2016}
Zhilin Yang, William Cohen, and Ruslan Salakhudinov.
\newblock Revisiting semi-supervised learning with graph embeddings.
\newblock In {\em International conference on machine learning}, pages 40--48.
  PMLR, 2016.

\bibitem{pei2020geomgcn}
Hongbin Pei, Bingzhe Wei, Kevin Chen-Chuan Chang, Yu~Lei, and Bo~Yang.
\newblock Geom-gcn: Geometric graph convolutional networks.
\newblock {\em arXiv preprint arXiv:2002.05287}, 2020.

\bibitem{Bogo:CVPR:2014}
Federica Bogo, Javier Romero, Matthew Loper, and Michael~J. Black.
\newblock {FAUST}: Dataset and evaluation for {3D} mesh registration.
\newblock In {\em Proceedings IEEE Conf. on Computer Vision and Pattern
  Recognition (CVPR)}, pages 3794 --3801, Columbus, Ohio, USA, June 2014.

\end{thebibliography}

\end{document}